\def\year{2020}\relax
\documentclass[letterpaper]{article} % DO NOT CHANGE THIS
\usepackage{aaai20}  % DO NOT CHANGE THIS
\usepackage{times}  % DO NOT CHANGE THIS
\usepackage{helvet} % DO NOT CHANGE THIS
\usepackage{courier}  % DO NOT CHANGE THIS
\usepackage[hyphens]{url}  % DO NOT CHANGE THIS
\usepackage{graphicx} % DO NOT CHANGE THIS
\usepackage{graphicx}  % DO NOT CHANGE THIS
\usepackage{amsmath,amssymb}
\usepackage{multirow}
\usepackage{algorithm}
\usepackage{algorithmicx}
\usepackage{algpseudocode}
\usepackage{enumitem}
\usepackage{booktabs}

\title{Factorized Inference in Deep Markov Models \\ for Incomplete Multimodal Time Series}
\author{Tan Zhi-Xuan,\textsuperscript{\rm 1, \rm 2}
Harold Soh,\textsuperscript{\rm 3}
Desmond C. Ong\textsuperscript{\rm 1, \rm 4}\\
\textsuperscript{\rm 1}{A*STAR Artificial Intelligence Initiative, A*STAR, Singapore}\\
\textsuperscript{\rm 2}{Department of Electrical Engineering and Computer Science, MIT}\\
\textsuperscript{\rm 3}{Department of Computer Science, National University of Singapore}\\
\textsuperscript{\rm 4}{Department of Information Systems and Analytics, National University of Singapore}\\
xuan@mit.edu, harold@comp.nus.edu.sg, 
dco@comp.nus.edu.sg}

\begin{document}

\maketitle

\begin{abstract}
  Integrating deep learning with latent state space models has the potential to yield temporal models that are powerful, yet tractable and interpretable. Unfortunately, current models are not designed to handle missing data or multiple data modalities, which are both prevalent in real-world data. In this work, we introduce a factorized inference method for Multimodal Deep Markov Models (MDMMs), allowing us to filter and smooth in the presence of missing data, while also performing uncertainty-aware multimodal fusion. We derive this method by factorizing the posterior $p(z|x)$ for non-linear state space models, and develop a \emph{variational backward-forward algorithm} for inference. Because our method handles incompleteness over both time and modalities, it is capable of interpolation, extrapolation, conditional generation, label prediction, and weakly supervised learning of multimodal time series. We demonstrate these capabilities on both synthetic and real-world multimodal data under high levels of data deletion. Our method performs well even with more than 50\% missing data, and outperforms existing deep approaches to inference in latent time series.
\end{abstract}

\section{Introduction}

Virtually all sensory data that humans and autonomous systems receive can be thought of as multimodal time series---multiple sensors each provide streams of information about the surrounding environment, and intelligent systems have to integrate these information streams to form a coherent yet dynamic model of the world. These time series are often asynchronously or irregularly sampled, with many time-points having missing or incomplete data. Classical time series algorithms, such as the Kalman filter, are robust to such incompleteness: they are able to infer the state of the world, but only in linear regimes \cite{kalman1960new}. On the other hand, human intelligence is robust \emph{and} complex: we infer complex quantities with \emph{nonlinear dynamics}, even from incomplete temporal observations of multiple modalities---for example, intended motion from both eye gaze and arm movements \cite{dragan2013legibility}, or desires and emotions from actions, facial expressions, speech, and language \cite{baker2017rational,ong2015affective}.

There has been a proliferation of attempts
to learn these nonlinear dynamics by integrating deep learning with traditional probabilistic approaches, such as Hidden Markov Models (HMMs) and latent dynamical systems.
Most approaches do this by adding random latent variables to each time step in an RNN \cite{chung2015recurrent,bayer2014learning,fabius2014variational,fraccaro2016sequential}. Other authors begin with latent sequence models as a basis, then develop deep parameterizations and inference techniques for these models \cite{krishnan2017structured,archer2015black,johnson2016composing,karl2016deep,doerr2018probabilistic,lin2018variational,chen2018neural}. Most relevant to our work are the Deep Markov Models (DMMs) proposed by \citeauthor{krishnan2017structured} (\citeyear{krishnan2017structured}), a generalization of HMMs and Gaussian State Space Models where the transition and emission distributions are parameterized by deep networks.

Unfortunately, none of the approaches thus far are designed to handle inference with both missing data and multiple modalities. Instead, most approaches rely upon RNN inference networks \cite{krishnan2017structured,karl2016deep,che2018multivariate}, which can only handle missing data using ad-hoc approaches such as zero-masking \cite{lipton2016directly}, update-skipping \cite{krishnan2017structured}, temporal gating mechanisms \cite{neil2016phased,che2018multivariate}, or providing time stamps as inputs \cite{chen2018neural}, none of which have intuitive probabilistic interpretations.
Of the approaches that do not rely on RNNs, Fraccaro \textit{et al.} handle missing data by assuming linear latent dynamics \cite{fraccaro2017disentangled}, while Johnson \textit{et al.} \cite{johnson2016composing} and Lin \textit{et al.} \cite{lin2018variational} use hybrid message passing inference that is theoretically capable of marginalizing out missing data. However, these methods are unable to learn nonlinear transition dynamics, nor do they handle multiple modalities.

\begin{figure*}[t]
    \centering
    \includegraphics[width=.85\textwidth]{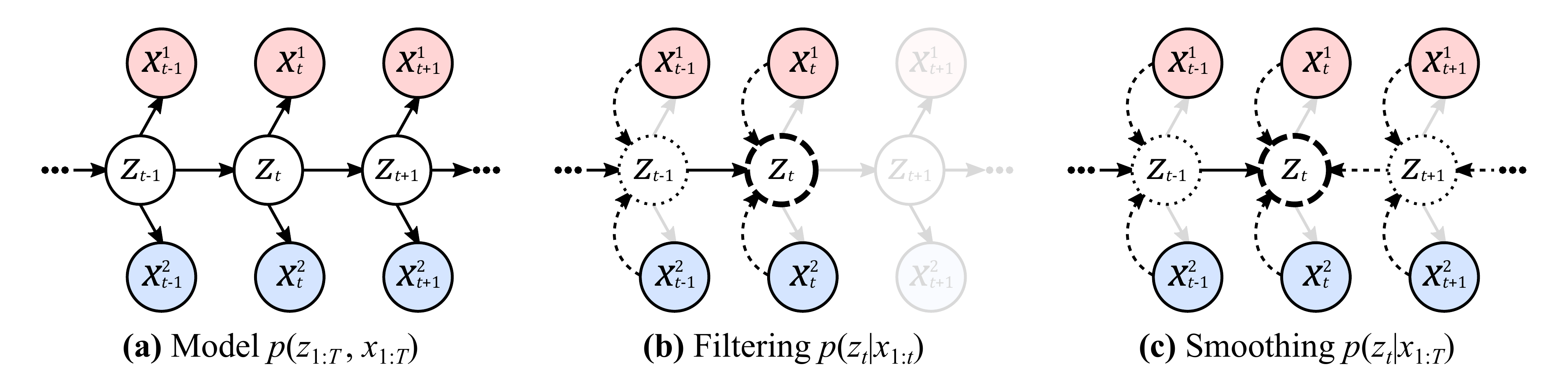}
    \caption{(a) A Multimodal Deep Markov Model (MDMM) with $M=2$ modalities. Observations (filled) are generated from unobserved latent states (unfilled). (b) Filtering infers the current latent state $z_t$ (bold dashed outline) given all observations up to $t$ (solid outlines), and marginalizes (dotted outline) over past latent states. (c) Smoothing infers $z_t$ given past, present, and future observations, and marginalizes over both past and future latent states.} % In (b) and (c), dashed curved arrows correspond to learned inference networks $q(z_t|x_t^m)$, solid rightward arrows correspond to the forward transition dynamics $p(z_t|z_{t-1})$, and dashed leftward arrows correspond to the backward transition dynamics $p(z_t|z_{t+1})$.}
    \label{fig:diagrams}
\end{figure*}

We address these limitations---the inability to handle missing data, over multiple modalities, and with nonlinear dynamics---by introducing a multimodal generalization of Deep Markov Models, as well as a factorized inference method for such models that handles missing time points and modalities by design. Our method allows for both filtering and smoothing given incomplete time series, while also performing uncertainty-aware multimodal fusion \textit{\`a la} the Multimodal Variational Autoencoder (MVAE) \cite{wu2018multimodal}. Because our method handles incompleteness over both time and modalities, it is capable of (i) interpolation, (ii) forward / backward extrapolation, and (iii) conditional generation of one modality from another, including label prediction. It also enables (iv) weakly supervised learning from incomplete multimodal data. We demonstrate these capabilities on both a synthetic dataset of noisy bidirectional spirals, as well as a real world dataset of labelled human actions. Our experiments show that our method learns and performs excellently on each of these tasks, while outperforming state-of-the-art inference methods that rely upon RNNs.

\section{Methods}

We introduce Multimodal Deep Markov Models (MDMMs) as a generalization of Krishnan \textit{et al.}'s  Deep Markov Models (DMMs)  \cite{krishnan2017structured}.
In a MDMM (Figure \ref{fig:diagrams}a), we model multiple sequences of observations, each of which is conditionally independent of the other sequences given the latent state variables. Each observation sequence corresponds to a particular data or sensor modality (e.g. video, audio, labels), and may be missing when other modalities are present. An MDMM can thus be seen as a sequential version of the MVAE \cite{wu2018multimodal}.

Formally, let $z_t$ and $x_t^m$ respectively denote the latent state and observation for modality $m$ at time $t$. An MDMM with $M$ modalities is then defined by the transition and emission distributions:
\begin{alignat}{3}
& z_t &&\sim \mathcal{N}(\mu_\theta(z_{t-1}), \Sigma_\theta(z_{t-1})) \quad &&(\text{Transition}) \\
& x_t^m &&\sim \Pi(\kappa_\theta^m(z_t)) \quad &&(\text{Emission})
\label{eqn:mdmm}
\end{alignat}
Here, $\mathcal{N}$ is the Gaussian distribution, and $\Pi$ is some arbitrary emission distribution. The distribution parameters $\mu_\theta$, $\Sigma_\theta$ and $\kappa^m_\theta$ are functions of either $z_{t-1}$ or $z_t$. We learn these functions as neural networks with weights $\theta$. We also use $z_{{t_1}:{t_2}}$ to denote the time-series of $z$ from $t_1$ to $t_2$, and $x_{t_1:t_2}^{{m_1}:{m_2}}$ to denote the corresponding observations from modalities $m_1$ to $m_2$. We omit the modality superscripts when all modalities are present (i.e., $x_t \equiv x_t^{1:M}$).

We want to jointly learn the parameters $\theta$ of the generative model $p_\theta(z_{1:T}, x_{1:T})=p_\theta(x_{1:T}|z_{1:T}) p_\theta(z_{1:T})$ and the parameters $\phi$ of a variational posterior $q_\phi(z_{1:T}|x_{1:T})$ which approximates the true (intractable) posterior $p_\theta(z_{1:T}|x_{1:T})$.
To do so, we maximize a lower bound on the log marginal likelihood $L(x;\theta,\phi) \leq p_\theta(x_{1:T})$, also known as the evidence lower bound (ELBO):
\begin{align}
L(x;\theta,\phi) &=  \mathbb{E}_{q_\phi(z_{1:T}|x_{1:T})} [ \log p_\theta(x_{1:T}|z_{1:T}) ] \\
&- \mathbb{E}_{q_\phi(z_{1:T}|x_{1:T})}  [ \text{KL}(q_\phi(z_{1:T}|x_{1:T}) || p_\theta(z_{1:T})) ] \nonumber
\label{eqn:elbo}
\end{align}

In practice, we can maximize the ELBO with respect to $\theta$ and $\phi$ via gradient ascent with stochastic backpropagation \cite{kingma2014auto,rezende2014stochastic}. Doing so requires sampling from the variational posterior $q_\phi(z_{1:T}|x_{1:T})$.
In the following sections, we derive a variational posterior that factorizes over time-steps and modalities, allowing us to tractably infer the latent states $z_{1:T}$ even when data is missing.

\subsection{Factorized Posterior Distributions}

In latent sequence models such as MDMMs, we often want to perform several kinds of inferences over the latent states. The most common of such latent state inferences are:

\begin{description}[noitemsep]
    \item[Filtering] Inferring $z_t$ given past observations $x_{1:t}$.
    \item[Smoothing] Inferring some $z_t$ given all observations $x_{1:T}$.
    \item[Sequencing] Inferring the sequence $z_{1:T}$ from $x_{1:T}$
\end{description}

Most architectures that combine deep learning with state space models focus upon filtering \cite{fabius2014variational,chung2015recurrent,hafner2018learning,buesing2018learning}, while Krishnan \textit{et al.} optimize their DMM for sequencing \cite{krishnan2017structured}. One of our contributions is to demonstrate that we can learn the filtering, smoothing, and sequencing distributions within the same framework, because they all share similar factorizations (see Figures \ref{fig:diagrams}b and \ref{fig:diagrams}c for the shared inference structure of filtering and smoothing). A further consequence of these factorizations is that we can naturally handle inferences given missing modalities or time-steps.

To demonstrate this similarity, we first factorize the sequencing distribution $p(z_{1:T}|x_{1:T})$ over time:
\begin{equation}
    p(z_{1:T}|x_{1:T}) = p(z_1|x_{1:T}) \textstyle\prod_{t=2}^T p(z_t|z_{t-1}, x_{t:T})
\label{eqn:sequencing-factorization}
\end{equation}
This factorization means that each latent state $z_t$ depends only on the previous latent state $z_{t-1}$, as well as all current and future observations $x_{t:T}$, and is implied by the graphical structure of the MDMM (Figure \ref{fig:diagrams}a). We term $p(z_t|z_{t-1}, x_{t:T})$ the \emph{conditional smoothing posterior}, because it is the posterior that corresponds to the \emph{conditional prior} $p(z_t|z_{t-1})$ on the latent space, and because it combines information from both past and future (hence `smoothing').

Given one or more modalities, we can show that the conditional smoothing posterior $p(z_t|z_{t-1}, x_{t:T})$, the \emph{backward} filtering distribution $p(z_t|x_{t:T})$, and the smoothing distribution $p(z_t|x_{1:T})$ all factorize almost identically:
\begin{align}
    &\textit{Backward Filtering} \label{eqn:bwd-filter-factor}  \\
    &p(z_t|x_{t:T}) \propto p(z_t|x_{t+1:T}) \left[ \textstyle\prod_{m} \frac{p(z_t|x^m_t)}{p(z_t)} \right] \nonumber
\\
    &\textit{Forward Smoothing} \label{eqn:fwd-smooth-factor}  \\ 
    &p(z_t|x_{1:T}) \propto p(z_t|x_{t+1:T}) \left[ \textstyle\prod_{m} \frac{p(z_t|x^m_t)}{p(z_t)} \right] \textstyle\frac{p(z_t|x_{1:t-1})}{p(z_t)} \nonumber
\\
    &\textit{Conditional Smoothing Posterior} \label{eqn:cond-smooth-factor} \\
    &p(z_t|z_{t-1}, x_{t:T}) \propto  p(z_t|x_{t+1:T}) \left[ \textstyle\prod_{m} \frac{p(z_t|x^m_t)}{p(z_t)} \right] \textstyle\frac{p(z_t|z_{t-1})}{p(z_t)} \nonumber 
\end{align}

Equations \ref{eqn:bwd-filter-factor}--\ref{eqn:cond-smooth-factor} show that each distribution can be decomposed into (i) its dependence on future observations, $p(z_t|x_{t+1:T})$, (ii) its dependence on each modality $m$ in the present, $p(z_t|x^m_t)$, and, excluding filtering, (iii) its dependence on the past $p(z_t|z_{t-1})$ or $p(z_t|x_{1:t-1})$. Their shared structure is due to the conditional independence of $x_{t:T}$ given $z_t$ from all prior observations or latent states. Here we show only the derivation for Equation \ref{eqn:cond-smooth-factor}, because the others follow by either dropping $z_{t-1}$ (Equation \ref{eqn:bwd-filter-factor}), or replacing $z_{t-1}$ with $x_{1:t-1}$ (Equation \ref{eqn:fwd-smooth-factor}):
\begin{align*}
&p(z_t|z_{t-1}, x_{t:T}^{1:M})
\\
&=
p(x^{1:M}_{t+1:T}|z_t)
p(x^{1:M}_t|z_t)
\textstyle\frac{p(z_t|z_{t-1})}
{p(x^{1:M}_{t:T}|z_{t-1})}
 \\
&\propto
p(x^{1:M}_{t+1:T}|z_t)
\left[ \textstyle\prod_{m=1}^M p(x^m_t|z_t) \right]
p(z_t|z_{t-1})
 \\
&=
\textstyle\frac{p(z_t|x^{1:M}_{t+1:T}) p(x^{1:M}_{t+1:T})}{p(z_t)}
\left[ \textstyle\prod_{m=1}^M \textstyle\frac{p(z_t|x^m_t) p(x^m_t)}{p(z_t)} \right]
p(z_t|z_{t-1})\\
&\propto
p(z_t|x^{1:M}_{t+1:T})
\left[ \textstyle\prod_{m=1}^M \textstyle\frac{p(z_t|x^m_t)}{p(z_t)} \right]
\textstyle\frac{p(z_t|z_{t-1})}{p(z_t)}
\end{align*}

The factorizations in Equations \ref{eqn:bwd-filter-factor}--\ref{eqn:cond-smooth-factor} lead to several useful insights. First, they show that any missing modalities $\overline m \in [1,M]$ at time $t$ can simply be left out of the product over modalities, leaving us with distributions that correctly condition on only the modalities $[1,M]\setminus \{\overline m\}$ that are present. Second, they suggest that we can compute all three distributions if we can approximate the dependence on the future, $q(z_t|x_{t+1:T}) \simeq p(z_t|x_{t+1:T})$, learn approximate posteriors $q(z_t|x^m_t) \simeq p(z_t|x^m_t)$ for each modality $m$, and know the model dynamics $p(z_t)$, $p(z_t|z_{t-1})$.

\subsection{Multimodal Fusion via Product of Gaussians}

However, there are a few obstacles to performing tractable computation of Equations \ref{eqn:bwd-filter-factor}--\ref{eqn:cond-smooth-factor}. One obstacle is that it is not tractable to compute the product of generic probability distributions. To address this, we adopt the approach used for the MVAE \cite{wu2018multimodal}, making the assumption that each term in Equations \ref{eqn:bwd-filter-factor}--\ref{eqn:cond-smooth-factor} is Gaussian. If each distribution is Gaussian, then their products or quotients are also Gaussian and can be computed in closed form. Since this result is well-known, we state it in the supplement (see  \citeauthor{wu2018multimodal} \shortcite{wu2018multimodal} for a proof).

This Product-of-Gaussians approach has the added benefit that the output distribution is dominated by the input Gaussian terms with lower variance (higher precision), thereby fusing information in a way that gives more weight to higher-certainty inputs \cite{cao2014generalized,ong2015affective}. This automatically balances the information provided by each modality $m$, depending on whether $p(z_t|x^m_t)$ is high or low certainty, as well as the information provided from the past and future through $p(z_t|z_{t-1})$ and $p(z_t|x_{t+1:T})$, thereby performing multimodal temporal fusion in a manner that is uncertainty-aware.

\subsection{Approximate Filtering with Missing Data}

Another obstacle to computing Equations \ref{eqn:bwd-filter-factor}--\ref{eqn:cond-smooth-factor} is the dependence on future observations, $p(z_t|x_{t+1:T})$, which does not admit further factorization, and hence does not readily handle missing data among those future observations. Other approaches to approximating this dependence on the future rely on RNNs as recognition models \cite{krishnan2017structured,che2018multivariate}, but these are not designed to work with missing data.

To address this obstacle in a more principled manner, our insight was that
$p(z_t|x_{t+1:T})$ is the expectation of $p(z_t|z_{t+1})$ under the backwards filtering distribution, $p(z_{t+1}|x_{t+1:T})$:
\begin{equation}
p(z_t|x_{t+1:T}) = \mathbb{E}_{p(z_{t+1}|x_{t+1:T})} \left[p(z_t|z_{t+1})\right]
\label{eqn:future-approx}
\end{equation}
For tractable approximation of this expectation, we use an approach similar to assumed density filtering \cite{huber2011semi}. We assume both $p(z_t|x_{t+1:T})$ and $p(z_t|z_{t+1})$ to be multivariate Gaussian with diagonal covariance, and sample the parameters $\mu$, $\Sigma$ of $p(z_t|z_{t+1})$ under $p(z_{t+1}|x_{t+1:T})$. After drawing $K$ samples, we approximate the parameters of $p(z_t|x_{t+1:T})$ via empirical moment-matching:
\begin{align}
    \hat\mu_{z_t|x_{t+1:T}} &= \textstyle\frac 1 K \textstyle\sum_{k=1}^K \mu_{k} \\\
    \hat\Sigma_{z_t|x_{t+1:T}} &= \textstyle\frac 1 K \textstyle\sum_{k=1}^K \left(\Sigma_{k} + \mu^2_{k}\right) - \hat\mu^2_{z_t|x_{t+1:T}}
\end{align}
Approximating $p(z_t|x_{t+1:T})$ by $p(z_t|z_{t+1})$ led us to three important insights. First, by substituting the expectation from Equation \ref{eqn:future-approx} into Equation \ref{eqn:bwd-filter-factor}, the backward filtering distribution becomes:
\begin{equation}
    p(z_t|x_{t:T}) \propto
\underset{{p(z_{t+1}|x_{t+1:T})}}{\mathbb{E}  \left[p(z_t|z_{t+1})\right]}
\left[ \textstyle\prod_{m=1}^M \frac{p(z_t|x^m_t)}{p(z_t)} \right]
\label{eqn:bwd-filter-recursion}
\end{equation}
In other words, by sampling under the filtering distribution for time $t+1$, $p(z_{t+1}|x_{t+1:T})$, we can compute the filtering distribution for time $t$, $p(z_t|x_{t:T})$. We can thus recursively compute $p(z_t|x_{t:T})$ backwards in time, starting from $t=T$.

Second, once we can perform filtering backwards in time, we can use this to approximate $p(z_t|x_{t+1:T})$ in the smoothing distribution (Equation \ref{eqn:fwd-smooth-factor}) and the conditional smoothing posterior (Equation \ref{eqn:cond-smooth-factor}). Backward filtering hence allows us to approximate both smoothing and sequencing.

Third, this approach removes the explicit dependence on all future observations $x_{t+1:T}$, allowing us to handle missing data. Suppose the data points $X_\nexists = \{x_{t_i}^{m_i}\}$ are missing, where $t_i$ and $m_i$ are the time-step and modality of the $i$th missing point respectively. Rather than directly compute the dependence on an incomplete set of future observations, $p(z_t|x_{t+1:T} \setminus X_\nexists)$, we can instead sample $z_{t+1}$ under the filtering distribution conditioned on incomplete observations, $p(z_{t+1}|x_{t+1:T} \setminus X_\nexists)$, and then compute $p(z_t|z_{t+1})$ given the sampled $z_{t+1}$, thereby approximating $p(z_t|x_{t+1:T} \setminus X_\nexists)$.

\subsection{Backward-Forward Variational Inference}

We now introduce factorized variational approximations of Equations \ref{eqn:bwd-filter-factor}--\ref{eqn:cond-smooth-factor}. We replace the true posteriors $p(z_t|x^m_t)$ with variational approximations $q(z_t|x^m_t){:=}\tilde q(z_t|x^m_t) p(z_t)$, where $\tilde q(z_t|x^m_t)$ is parameterized by a (time-invariant) neural network for each modality $m$. As in the MVAE, we learn the Gaussian quotients $\tilde q(z_t|x^m_t){:=}q(z_t|x^m_t) / p(z_t)$ directly, so as to avoid the constraint required for ensuring a quotient of Gaussians is well-defined.  We also parameterize the transition dynamics $p(z_t|z_{t-1})$ and $p(z_t|z_{t+1})$ using neural networks for the quotient distributions.
This gives the following approximations:
\begin{align}
    &\textit{Backward Filtering}     \label{eqn:bwd-filter-approx}\\
    &q(z_t|x_{t:T}) \propto  \mathbb{E}_{\gets} \left[ p(z_t|z_{t+1})\right]
    \textstyle\prod_m \tilde{q}(z_t|x^m_t) \nonumber \\
    &\textit{Forward Smoothing} \label{eqn:fwd-smooth-approx} \\
    &q(z_t|x_{1:T}) \propto  \mathbb{E}_{\gets} \left[ p(z_t|z_{t+1})\right]
    \textstyle\prod_m \tilde{q}(z_t|x^m_t)\   \textstyle\frac{\mathbb{E}_{\to} \left[p(z_t|z_{t-1})\right]}{p(z_t)} \nonumber \\
    &\textit{Conditional Smoothing Posterior}     \label{eqn:cond-smooth-approx} \\
    &q(z_t|z_{t-1}, x_{t:T}) \propto  \mathbb{E}_{\gets} \left[ p(z_t|z_{t+1})\right]
    \textstyle\prod_m \tilde{q}(z_t|x^m_t)\   \textstyle\frac{p(z_t|z_{t-1})}{p(z_t)} \nonumber
\end{align}

Here, $\mathbb{E}_{\gets}$ is shorthand for the expectation under the approximate backward filtering distribution $q(z_{t+1}|x_{t+1:T})$, while  $\mathbb{E}_{\to}$ is the expectation under the forward smoothing distribution $q(z_{t-1}|x_{1:T})$.

To calculate the backward filtering distribution $q(z_t|x_{t:T})$, we introduce a \emph{variational backward algorithm} (Algorithm \ref{alg:bwd-filter}) to recursively compute Equation \ref{eqn:bwd-filter-approx} for all time-steps $t$ in a single pass. Note that simply by reversing time in Algorithm \ref{alg:bwd-filter}, this gives us a \emph{variational forward algorithm} that computes the forward filtering distribution $q(z_t|x_{1:t})$.

Unlike filtering, smoothing and sequencing require information from both past ($p(z_t|z_{t-1})$) and future ($p(z_t|z_{t+1})$). This motivates a \emph{variational backward-forward algorithm} (Algorithm \ref{alg:fwd-smooth}) for smoothing and sequencing. Algorithm \ref{alg:fwd-smooth} first uses Algorithm \ref{alg:bwd-filter} as a backward pass, then performs a forward pass to propagate information from past to future. Algorithm \ref{alg:fwd-smooth} also requires knowing $p(z_t)$ for each $t$. While this can be computed by sampling in the forward pass, we avoid the instability
(of sampling $T$ successive latents with no observations)
by instead assuming $p(z_t)$ is constant with time, i.e., the MDMM is stationary when nothing is observed. During training, we add $\text{KL}(p(z_t)||\mathbb{E}_{z_{t-1}} p(z_t|z_{t-1}))$ and $\text{KL}(p(z_t)||\mathbb{E}_{z_{t+1}} p(z_t|z_{t+1}))$ to the loss to ensure that the transition dynamics obey this assumption.

\begin{algorithm}[t]
\caption{A variational backward algorithm for approximate backward filtering.}
\label{alg:bwd-filter}
\begin{algorithmic}
\footnotesize
\Function{BackwardFilter}{$x_{1:T}$, $K$}
\State Initialize $q(z_t|x_{T+1:T}) \gets p(z_T)$
\For{$t=T$ to $1$}
    \State Let $\mathcal{M} \subseteq [1,M]$ be the observed modalities at $t$
    \State $q(z_t|x_{t:T}) \gets
    q(z_t|x_{t+1:T}) \prod_\mathcal{M} \tilde{q}(z_t|x^m_t)$
    \State Sample $K$ particles $z_t^k \sim q(z_t|x_{t:T})$ for $k\in[1,K]$
    \State Compute $p(z_{t-1}|z_t^k)$ for each particle $z_t^k$
    \State $q(z_{t-1}|x_{t:T}) \gets \frac 1 K \sum_{k=1}^K p(z_{t-1}|z_t^k)$
\EndFor
\State \Return  $\{ q(z_t|x_{t:T})$, $q(z_t|x_{t+1:T}) \text{ for }t \in [1,T]\}$
\EndFunction
\end{algorithmic}
\end{algorithm}

\begin{algorithm}[t]
\caption{A variational backward-forward algorithm for approximate forward smoothing.}
\label{alg:fwd-smooth}
\begin{algorithmic}
\footnotesize
\Function{ForwardSmooth}{$x_{1:T}$, $K_b$, $K_f$}
\State Initialize $\tilde{p}(z_t|x_{1:0}) \gets 1$
\State Collect $q(z_t|x_{t+1:T})$ from \Call{BackwardFilter}{$x_{1:T}$, $K_b$}
\For{$t=1$ to $T$}
    \State Let $\mathcal{M} \subseteq [1,M]$ be the observed modalities at $t$
    \State $q(z_t|x_{1:T}) \gets
    q(z_t|x_{t+1:T}) \prod_\mathcal{M} [\tilde{q}(z_t|x^m_t)] \frac{q(z_t|x_{1:t-1})}{p(z_t)}$
    \State Sample $K_f$ particles $z_t \sim q(z_t|x_{1:T})$  for $k\in[1,K_f]$
    \State Compute $p(z_{t+1}|z_t^k)$ for each particle $z_t^k$
    \State $q(z_{t+1}|x_{1:t}) \gets \frac{1}{K_f} \sum_{k=1}^{K_f} p(z_{t+1}|z_t^k)$
\EndFor
\State \Return $\{ q(z_t|x_{1:T})$, $q(z_t|x_{1:t-1}) \text{ for }t \in [1,T]\}$
\EndFunction
\end{algorithmic}
\end{algorithm}

While Algorithm \ref{alg:bwd-filter} approximates the filtering distribution $q(z_t|x_{t:T})$, by setting the number of particles $K=1$, it effectively computes the (backward) conditional filtering posterior $q(z_t|z_{t+1},x_t)$ and (backward) conditional prior $p(z_t|z_{t+1})$ for a randomly sampled latent sequence $z_{1:T}$. Similarly, while Algorithm \ref{alg:fwd-smooth} approximates smoothing by default, when $K_f=1$, it effectively computes the (forward) conditional smoothing posterior $q(z_t|z_{t-1},x_{t:T})$ and (forward) conditional prior $p(z_t|z_{t-1})$ for a random latent sequence $z_{1:T}$. These quantities are useful not only because they allow us to perform sequencing, but also because we can use them to compute the ELBO for both backward filtering and forward smoothing:
\begin{align}
    &L_\text{filter} = \textstyle\sum_{t=1}^T \big[\mathbb{E}_{q(z_t|x_{t:T})} \log p(x_t|z_t) - \nonumber \\
    &\mathbb{E}_{q(z_{t+1}|x_{t+1:T})} \text{KL}\big(q(z_t|z_{t+1},x_t) || p(z_t|z_{t+1})\big) \big] \\
    &L_\text{smooth} = \textstyle\sum_{t=1}^T \big[ \mathbb{E}_{q(z_t|x_{1:T})} \log p(x_t|z_t) - \nonumber \\
    &\mathbb{E}_{q(z_{t-1}|x_{1:T})} \text{KL}\big(q(z_t|z_{t-1},x_{t:T}) || p(z_t|z_{t-1})\big) \big]
\end{align}

\begin{table*}[htb]
\center
\begin{tabular}{@{}lcccccc@{}}
\toprule
\textbf{Method} &
 \textit{Recon.} & \textit{Drop Half} & \textit{Fwd. Extra.} & \textit{Bwd. Extra.} & \textit{Cond. Gen.} & \textit{Label Pred.} \\ \hline
 & \multicolumn{6}{c}{\textbf{Spirals Dataset: MSE (SD)} \rule{0pt}{2.6ex}} \\
 \cmidrule(l){2-7}
BFVI (ours)  & \textbf{0.02 (0.01)} & \textbf{0.04 (0.01)} & 0.12 (0.10) & 0.07 (0.03) & \textbf{0.26 (0.26)} & -- \\
F-Mask & \textbf{0.02 (0.01)} & 0.06 (0.02) & \textbf{0.10 (0.08)} & 0.18 (0.07) & 1.37 (1.39) & --\\
F-Skip & 0.04 (0.01) & 0.10 (0.05) & 0.13 (0.11) & 0.19 (0.06) & 1.51 (1.54) & --\\
B-Mask & \textbf{0.02 (0.01)} & \textbf{0.04 (0.01)} & 0.18 (0.14) & \textbf{0.04 (0.01)} & 1.25 (1.23) & --\\
B-Skip & 0.05 (0.01) & 0.19 (0.05) & 0.32 (0.22) & 0.37 (0.15) & 1.64 (1.51) & --\\ \bottomrule
 & \multicolumn{6}{c}{\textbf{Weizmann Video Dataset: SSIM or Accuracy* (SD)} \rule{0pt}{2.6ex}}
  \\ \cmidrule(l){2-7}
BFVI (ours) & \textbf{.85 (.03)} & \textbf{.84 (.04)} & \textbf{.84 (.04)} & \textbf{.83 (.05)} & \textbf{.85 (.03)} & \textbf{.69 (.33)}* \\
F-Mask & .68 (.18) & .66 (.18) & .68 (.18) & .66 (.17) & .60 (.15) & .33 (.33)* \\
F-Skip & .70 (.12) & .68 (.14) & .70 (.12) & .67 (.16) & .63 (.12) & .21 (.26)* \\
B-Mask & .79 (.04) & .79 (.04) & .79 (.04) & .79 (.04) & .76 (.06) & .46 (.34)* \\
B-Skip & .80 (.04) & .79 (.04) & .80 (.04) & .79 (.04) & .74 (.08) & .29 (.37)* \\ \bottomrule
\end{tabular}
\caption{Evaluation metrics on both datasets across inference methods and tasks. Best performance per task (column) in bold. (Top) \textit{Spirals Dataset:} MSE (lower is better) per time-step between reconstructions and ground truth spirals. For scale, the average squared spiral radius is about 5 sq. units. (Bottom) \textit{Weizmann Video Dataset:} SSIM or label accuracy (higher is better) per time-step with respect to original videos. Means and Standard Deviations (SD) are across the test set.}
\label{tab:results}
\end{table*}

$L_\text{filter}$ is the filtering ELBO because it corresponds to a `backward filtering' variational posterior $q(z_{1:T}|x_{1:T}) = \prod_t q(z_t|z_{t+1}, x_t)$, where each $z_t$ is only inferred using the current observation $x_t$ and the future latent state $z_{t+1}$. $L_\text{smooth}$ is the smoothing ELBO because it corresponds to the correct factorization of the posterior in Equation \ref{eqn:sequencing-factorization}, where each term combines information from both past and future. Since $L_\text{smooth}$ corresponds to the correct factorization, it should theoretically be enough to minimize $L_\text{smooth}$ to learn good MDMM parameters $\theta, \phi$. However, in order to compute $L_\text{smooth}$, we must perform a backward pass which requires sampling under the backward filtering distribution. Hence, to accurately approximate $L_\text{smooth}$, the backward filtering distribution has to be reasonably accurate as well. This motivates learning the parameters $\theta, \phi$ by jointly maximizing the filtering and smoothing ELBOs as a weighted sum. We call this paradigm \textbf{backward-forward variational inference (BFVI)}, due to its use of variational posteriors for both backward filtering and forward smoothing.

\section{Experiments}

We compare \textbf{BFVI} against state-of-the-art RNN-based inference methods on two multimodal time series datasets over a range of inference tasks. \textbf{F-Mask} and \textbf{F-Skip} use forward RNNs (one per modality), using zero-masking and update skipping respectively to handle missing data. They are thus multimodal variants of the ST-L network in \cite{krishnan2017structured}, and similar to the variational RNN \cite{chung2015recurrent} and recurrent SSM \cite{hafner2018learning}. \textbf{B-Mask} and \textbf{B-Skip} use backward RNNs, with masking and skipping respectively, and correspond to the Deep Kalman Smoother in \cite{krishnan2017structured}. The underlying MDMM architecture is constant across inference methods. Architectural and training details can be found in the supplement. Code is available at \url{https://git.io/Jeoze}.

\subsection{Datasets}

\begin{description}[leftmargin=0pt]
\item[Noisy Spirals.] We synthesized a dataset of $1000$ noisy 2D spirals (600 train / 400 test) similar to Chen \textit{et al.} \cite{chen2018neural}, treating the $x$ and $y$ coordinates as two separate modalities. Spiral trajectories vary in direction (clockwise or counter-clockwise), size, and aspect ratio, and Gaussian noise is added to the observations. We used $5$ latent dimensions, and two-layer perceptrons for encoding $q(z_t|x_t^m)$ and decoding $p(x_t^m|z_t)$. For evaluation, we compute the mean squared error (MSE) per time step between the predicted trajectories and ground truth spirals.

\item[Weizmann Human Actions.] This is a video dataset of 9 people each performing 10 actions \cite{gorelick2007actions}. We converted it to a trimodal time series dataset by treating silhouette masks as an additional modality, and treating actions as per-frame labels, similar to \citeauthor{he2018probabilistic} (\citeyear{he2018probabilistic}). Each RGB frame was cropped to the central 128$\times$128 window and resized to 64$\times$64. We selected one person's videos as the test set, and the other 80 videos as the training set, allowing us to test action label prediction on an unseen person. We used $256$ latent dimensions, and convolutional / deconvolutional neural networks for encoding and decoding.  For evaluation, we compute the Structural Similarity (SSIM) between the input video frames and the reconstructed outputs.
\end{description}

\subsection{Inference Tasks}

We evaluated all methods on the following suite of temporal inference tasks for both datasets:
\textbf{reconstruction}: reconstruction given complete observations;
\textbf{drop half}: reconstruction after half of the inputs are randomly deleted;
\textbf{forward extrapolation}: predicting the last 25\% of a sequence when the rest is given; and \textbf{backward extrapolation}: inferring the first 25\% of a sequence when the rest is given.

When evaluating these tasks on the Weizmann dataset, we provided only video frames as input (i.e. with neither the silhouette masks nor the action labels), to test whether the methods were capable of unimodal inference after multimodal training.

We also tested cross-modal inference using the following \emph{conditional generation / label prediction} tasks: \textbf{conditional generation (Spirals)}: given $x$-  and initial 25\% of $y$-coordinates, generate rest of spiral;
\textbf{conditional generation (Weizmann)}: given the video frames, generate the silhouette masks; and
\textbf{label prediction (Weizmann)}: infer action labels given only video frames.

\begin{figure}[t]
    \centering
    \includegraphics[width=\columnwidth]{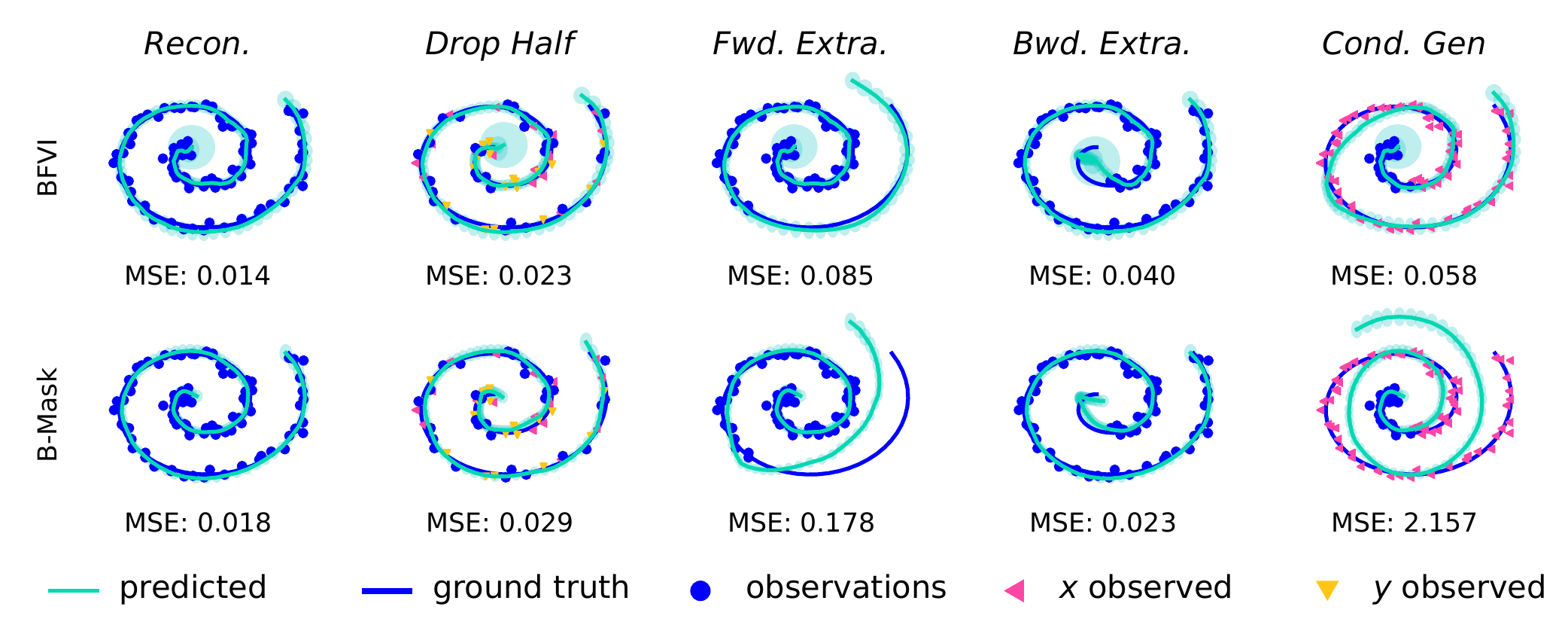}
    \caption{Reconstructions for all 5 spiral inference tasks for BFVI and the next best method, B-Mask. BFVI outperforms B-Mask significantly on both forward extrapolation and conditional generation.}
    \label{fig:spirals}
\end{figure}

\begin{figure}[htb]
    \centering
    \includegraphics[width=\columnwidth]{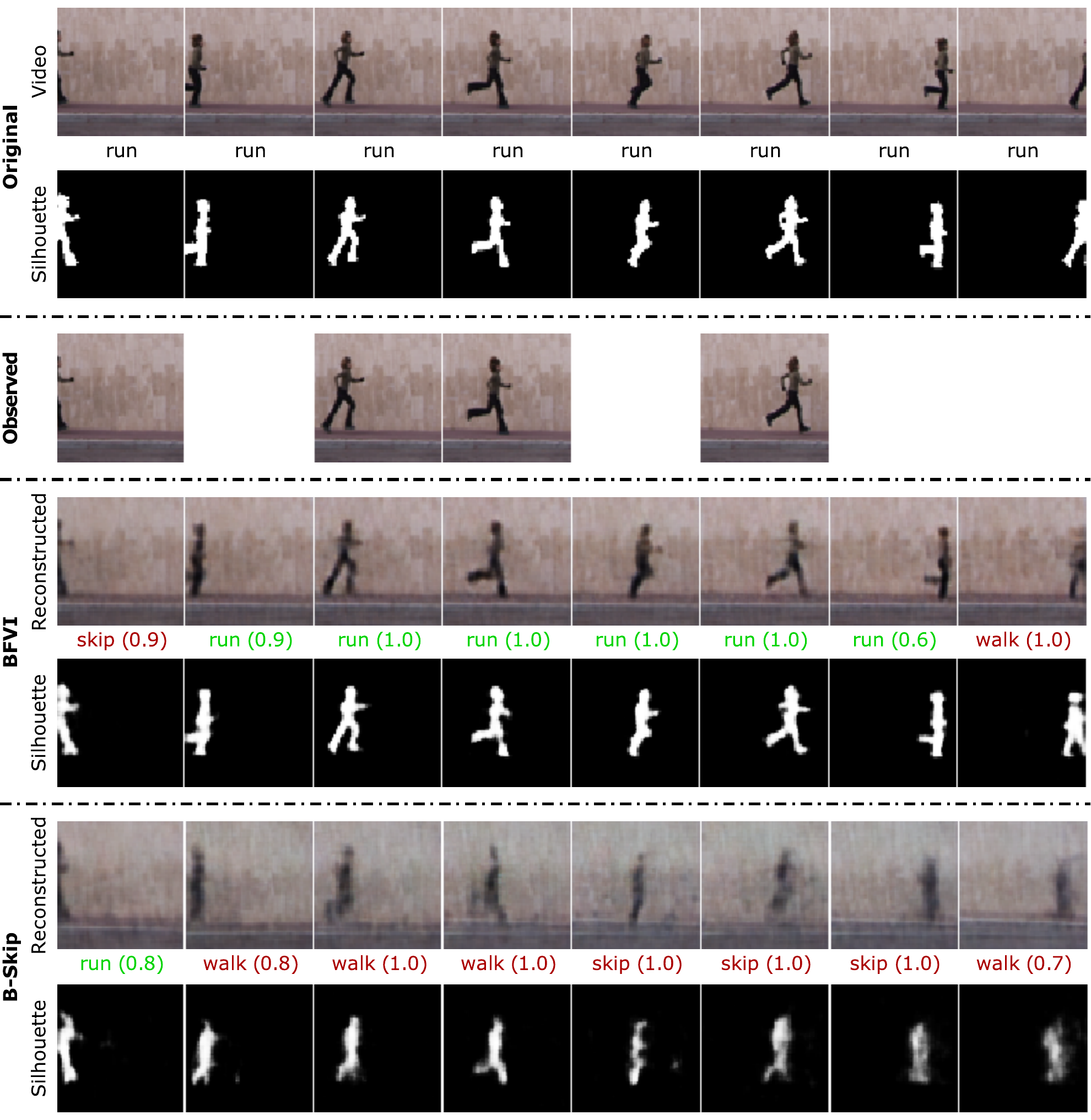}
    \caption{Snapshots of a `running' video and silhouette mask from the Weizmann dataset (rows 1--2), with half of the frames deleted at random (14 out of 28 frames), and neither action labels nor silhouettes provided as observations (row 3). BFVI reconstructs the video and a running silhouette, and also correctly predicts the action (rows 4--5). By contrast, B-Skip (the next best method) creates blurred and wispy reconstructions, wrongly predicts the action label, and vacillates between different possible action silhouettes over time (rows 6--7).}
    \label{fig:video}
\end{figure}

Table \ref{tab:results} shows the results for the inference tasks, while Figure \ref{fig:spirals} and \ref{fig:video} show sample reconstructions from the Spirals and Weizmann datasets respectively. On the Spirals dataset, BFVI achieves high performance on all tasks, whereas the RNN-based methods only perform well on a few.
In particular, all methods besides BFVI do poorly on the conditional generation task, which can be understood from the right-most column of Figure \ref{fig:spirals}. BFVI generates a spiral that matches the provided $x$-coordinates, while the next-best method, B-Mask, completes the trajectory with a plausible spiral, but ignores the $x$ observations entirely in the process.

On the more complex Weizmann video dataset, BFVI outperforms all other methods on every task, demonstrating both the power and flexibility of our approach. The RNN-based methods performed especially poorly on label prediction, and this was the case even on the training set (not shown in Table \ref{tab:results}). We suspect that this is because the RNN-based methods lack a principled approach to multimodal fusion, and hence fail to learn a latent space which captures the mutual information between action labels and images. In contrast, BFVI learns to both predict one modality from another, and to propagate information across time, as can be seen from the reconstruction and predictions in Figure \ref{fig:video}.

\subsection{Weakly Supervised Learning}

In addition to performing inference with missing data test time, we compared the various methods ability to learn with missing data at training time, amounting to a form of weakly supervised learning. We tested two forms of weakly supervised learning on the Spirals dataset, corresponding to different conditions of data incompleteness. The first was learning with data missing uniformly at random. This condition can arise when sensors are noisy or asynchronous. The second was learning with missing modalities, or semi-supervised learning, where a fraction of the sequences in the dataset only has a single modality present. This condition can arise when a sensor breaks down, or when the dataset is partially unlabelled by annotators. We also tested learning with missing modalities on the Weizmann dataset.

\begin{figure*}[htb]
  \centering
  \includegraphics[width=\textwidth]{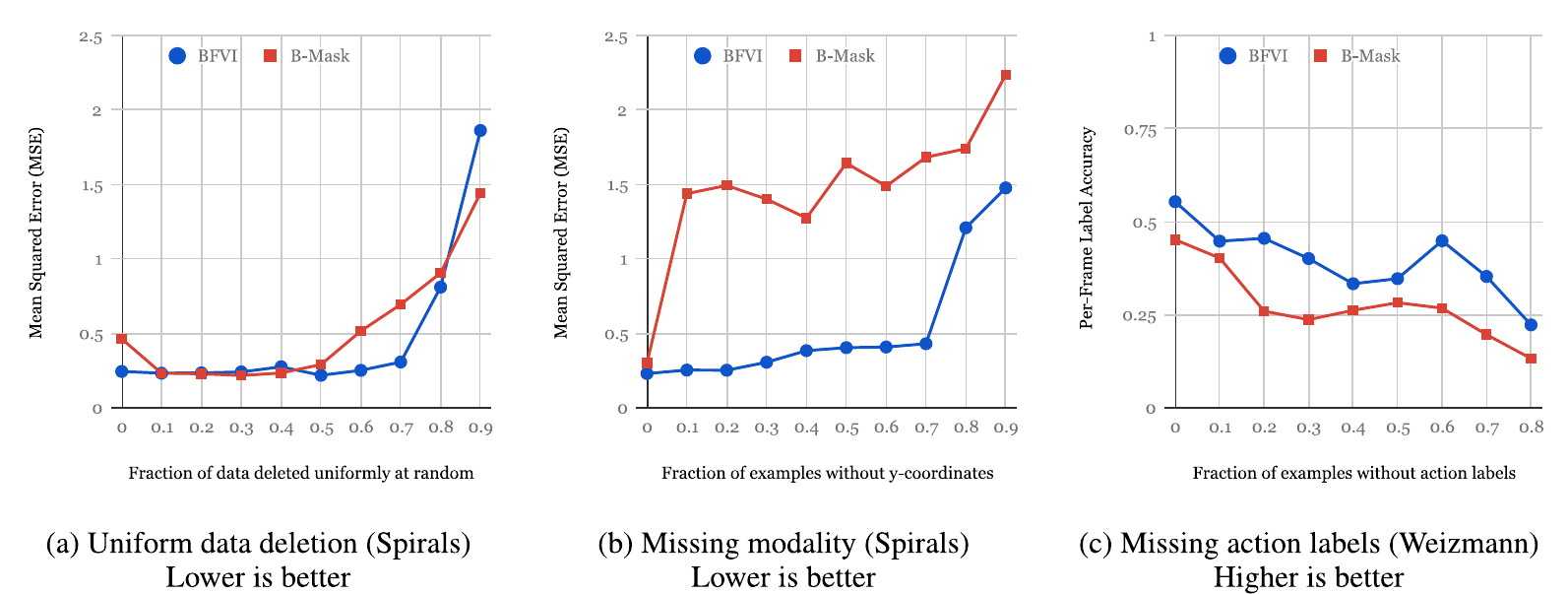}
  \caption{Learning curves under various forms of weak supervision. (a) Learning with randomly missing data on the Spirals dataset. (b) Semi-supervised learning on the Spirals dataset, where some sequences are entirely missing the y-coordinate. (c) Semi-supervised learning on the Weizmann dataset, where some sequences have no action labels.}
  \label{fig:weaksup}
\end{figure*}

Results for these experiments are shown in Figure \ref{fig:weaksup}, which compare BFVI's performance on increasing levels of missing data against the next best method, averaged across 10 trials. Our method (BFVI) performs well, maintaining good performance on the Spirals dataset even with 70\% uniform random deletion (Figure \ref{fig:weaksup}a) and 60\% uni-modal examples (Figure \ref{fig:weaksup}b), while degrading gracefully with increasing missingness. This is in contrast to B-Mask, which is barely able to learn when even $10\%$ of the spiral examples are uni-modal, and performs worse than BFVI on the Weizmann dataset at all levels of missing data (Figure \ref{fig:weaksup}c).

\section{Conclusion}

In this paper, we introduced backward-forward variational inference (BFVI) as a novel inference method for Multimodal Deep Markov Models. This method handles incomplete data via a factorized variational posterior, allowing us to easily marginalize over missing observations. Our method is thus capable of a large range of multimodal temporal inference tasks, which we demonstrate on both a synthetic dataset and a video dataset of human motions. The ability to handle missing data also enables applications in weakly supervised learning of labelled time series.
Given the abundance of multimodal time series data where missing data is the norm rather than the exception, our work holds great promise for many future applications.

\section{Acknowledgements}

This work was supported by the A*STAR Human-Centric Artificial Intelligence Programme (SERC SSF Project No. A1718g0048).

% \fontsize{9.5pt}{10.5pt} \selectfont
\bibliographystyle{aaai}
\bibliography{sources.bib}

\begin{thebibliography}{}

\bibitem[\protect\citeauthoryear{Archer \bgroup et al\mbox.\egroup
  }{2015}]{archer2015black}
Archer, E.; Park, I.~M.; Buesing, L.; Cunningham, J.; and Paninski, L.
\newblock 2015.
\newblock Black box variational inference for state space models.
\newblock {\em arXiv preprint arXiv:1511.07367}.

\bibitem[\protect\citeauthoryear{Baker \bgroup et al\mbox.\egroup
  }{2017}]{baker2017rational}
Baker, C.~L.; Jara-Ettinger, J.; Saxe, R.; and Tenenbaum, J.~B.
\newblock 2017.
\newblock Rational quantitative attribution of beliefs, desires and percepts in
  human mentalizing.
\newblock {\em Nature Human Behaviour} 1(4):0064.

\bibitem[\protect\citeauthoryear{Bayer and
  Osendorfer}{2014}]{bayer2014learning}
Bayer, J., and Osendorfer, C.
\newblock 2014.
\newblock Learning stochastic recurrent networks.
\newblock In {\em NIPS 2014 Workshop on Advances in Variational Inference}.

\bibitem[\protect\citeauthoryear{Buesing \bgroup et al\mbox.\egroup
  }{2018}]{buesing2018learning}
Buesing, L.; Weber, T.; Racaniere, S.; Eslami, S.; Rezende, D.; Reichert,
  D.~P.; Viola, F.; Besse, F.; Gregor, K.; Hassabis, D.; et~al.
\newblock 2018.
\newblock Learning and querying fast generative models for reinforcement
  learning.
\newblock {\em arXiv preprint arXiv:1802.03006}.

\bibitem[\protect\citeauthoryear{Cao and Fleet}{2014}]{cao2014generalized}
Cao, Y., and Fleet, D.~J.
\newblock 2014.
\newblock Generalized product of experts for automatic and principled fusion of
  gaussian process predictions.
\newblock In {\em Modern Nonparametrics 3: Automating the Learning Pipeline
  Workshop, NeurIPS 2014.}

\bibitem[\protect\citeauthoryear{Che \bgroup et al\mbox.\egroup
  }{2018}]{che2018multivariate}
Che, Z.; Purushotham, S.; Li, G.; Jiang, B.; and Liu, Y.
\newblock 2018.
\newblock Hierarchical deep generative models for multi-rate multivariate time
  series.
\newblock In Dy, J., and Krause, A., eds., {\em Proceedings of the 35th
  International Conference on Machine Learning}, volume~80 of {\em Proceedings
  of Machine Learning Research},  784--793.
\newblock Stockholmsmässan, Stockholm Sweden: PMLR.

\bibitem[\protect\citeauthoryear{Chen \bgroup et al\mbox.\egroup
  }{2018}]{chen2018neural}
Chen, T.~Q.; Rubanova, Y.; Bettencourt, J.; and Duvenaud, D.~K.
\newblock 2018.
\newblock Neural ordinary differential equations.
\newblock In {\em Advances in Neural Information Processing Systems},
  6571--6583.

\bibitem[\protect\citeauthoryear{Chung \bgroup et al\mbox.\egroup
  }{2015}]{chung2015recurrent}
Chung, J.; Kastner, K.; Dinh, L.; Goel, K.; Courville, A.~C.; and Bengio, Y.
\newblock 2015.
\newblock A recurrent latent variable model for sequential data.
\newblock In {\em Advances in neural information processing systems},
  2980--2988.

\bibitem[\protect\citeauthoryear{Doerr \bgroup et al\mbox.\egroup
  }{2018}]{doerr2018probabilistic}
Doerr, A.; Daniel, C.; Schiegg, M.; Duy, N.-T.; Schaal, S.; Toussaint, M.; and
  Sebastian, T.
\newblock 2018.
\newblock Probabilistic recurrent state-space models.
\newblock In Dy, J., and Krause, A., eds., {\em Proceedings of the 35th
  International Conference on Machine Learning}, volume~80 of {\em Proceedings
  of Machine Learning Research},  1280--1289.
\newblock Stockholmsmässan, Stockholm Sweden: PMLR.

\bibitem[\protect\citeauthoryear{Dragan, Lee, and
  Srinivasa}{2013}]{dragan2013legibility}
Dragan, A.~D.; Lee, K.~C.; and Srinivasa, S.~S.
\newblock 2013.
\newblock Legibility and predictability of robot motion.
\newblock In {\em Proceedings of the 8th ACM/IEEE international conference on
  Human-robot interaction},  301--308.
\newblock IEEE Press.

\bibitem[\protect\citeauthoryear{Fabius and van
  Amersfoort}{2014}]{fabius2014variational}
Fabius, O., and van Amersfoort, J.~R.
\newblock 2014.
\newblock Variational recurrent auto-encoders.
\newblock {\em arXiv preprint arXiv:1412.6581}.

\bibitem[\protect\citeauthoryear{Fraccaro \bgroup et al\mbox.\egroup
  }{2016}]{fraccaro2016sequential}
Fraccaro, M.; S{\o}nderby, S.~K.; Paquet, U.; and Winther, O.
\newblock 2016.
\newblock Sequential neural models with stochastic layers.
\newblock In {\em Advances in Neural Information Processing Systems},
  2199--2207.

\bibitem[\protect\citeauthoryear{Fraccaro \bgroup et al\mbox.\egroup
  }{2017}]{fraccaro2017disentangled}
Fraccaro, M.; Kamronn, S.; Paquet, U.; and Winther, O.
\newblock 2017.
\newblock A disentangled recognition and nonlinear dynamics model for
  unsupervised learning.
\newblock In {\em Advances in Neural Information Processing Systems},
  3601--3610.

\bibitem[\protect\citeauthoryear{Gorelick \bgroup et al\mbox.\egroup
  }{2007}]{gorelick2007actions}
Gorelick, L.; Blank, M.; Shechtman, E.; Irani, M.; and Basri, R.
\newblock 2007.
\newblock Actions as space-time shapes.
\newblock {\em IEEE transactions on pattern analysis and machine intelligence}
  29(12):2247--2253.

\bibitem[\protect\citeauthoryear{Hafner \bgroup et al\mbox.\egroup
  }{2018}]{hafner2018learning}
Hafner, D.; Lillicrap, T.; Fischer, I.; Villegas, R.; Ha, D.; Lee, H.; and
  Davidson, J.
\newblock 2018.
\newblock Learning latent dynamics for planning from pixels.
\newblock {\em arXiv preprint arXiv:1811.04551}.

\bibitem[\protect\citeauthoryear{He \bgroup et al\mbox.\egroup
  }{2018}]{he2018probabilistic}
He, J.; Lehrmann, A.; Marino, J.; Mori, G.; and Sigal, L.
\newblock 2018.
\newblock Probabilistic video generation using holistic attribute control.
\newblock In {\em Proceedings of the European Conference on Computer Vision
  (ECCV)},  452--467.

\bibitem[\protect\citeauthoryear{Huber, Beutler, and
  Hanebeck}{2011}]{huber2011semi}
Huber, M.~F.; Beutler, F.; and Hanebeck, U.~D.
\newblock 2011.
\newblock Semi-analytic gaussian assumed density filter.
\newblock In {\em Proceedings of the 2011 American Control Conference},
  3006--3011.
\newblock IEEE.

\bibitem[\protect\citeauthoryear{Johnson \bgroup et al\mbox.\egroup
  }{2016}]{johnson2016composing}
Johnson, M.; Duvenaud, D.~K.; Wiltschko, A.; Adams, R.~P.; and Datta, S.~R.
\newblock 2016.
\newblock Composing graphical models with neural networks for structured
  representations and fast inference.
\newblock In {\em Advances in Neural Information Processing Systems},
  2946--2954.

\bibitem[\protect\citeauthoryear{Kalman}{1960}]{kalman1960new}
Kalman, R.~E.
\newblock 1960.
\newblock A new approach to linear filtering and prediction problems.
\newblock {\em Journal of basic Engineering} 82(1):35--45.

\bibitem[\protect\citeauthoryear{Karl \bgroup et al\mbox.\egroup
  }{2016}]{karl2016deep}
Karl, M.; Soelch, M.; Bayer, J.; and van~der Smagt, P.
\newblock 2016.
\newblock Deep variational bayes filters: Unsupervised learning of state space
  models from raw data.
\newblock {\em arXiv preprint arXiv:1605.06432}.

\bibitem[\protect\citeauthoryear{Kingma and Welling}{2014}]{kingma2014auto}
Kingma, D.~P., and Welling, M.
\newblock 2014.
\newblock Auto-encoding variational bayes.
\newblock In {\em International Conference on Learning Representations}.

\bibitem[\protect\citeauthoryear{Krishnan, Shalit, and
  Sontag}{2017}]{krishnan2017structured}
Krishnan, R.~G.; Shalit, U.; and Sontag, D.
\newblock 2017.
\newblock Structured inference networks for nonlinear state space models.
\newblock In {\em Thirty-First AAAI Conference on Artificial Intelligence}.

\bibitem[\protect\citeauthoryear{Lin, Khan, and
  Hubacher}{2018}]{lin2018variational}
Lin, W.; Khan, M.~E.; and Hubacher, N.
\newblock 2018.
\newblock Variational message passing with structured inference networks.
\newblock In {\em International Conference on Learning Representations}.

\bibitem[\protect\citeauthoryear{Lipton, Kale, and
  Wetzel}{2016}]{lipton2016directly}
Lipton, Z.~C.; Kale, D.; and Wetzel, R.
\newblock 2016.
\newblock Directly modeling missing data in sequences with rnns: Improved
  classification of clinical time series.
\newblock In {\em Machine Learning for Healthcare Conference},  253--270.

\bibitem[\protect\citeauthoryear{Neil, Pfeiffer, and
  Liu}{2016}]{neil2016phased}
Neil, D.; Pfeiffer, M.; and Liu, S.-C.
\newblock 2016.
\newblock Phased lstm: Accelerating recurrent network training for long or
  event-based sequences.
\newblock In {\em Advances in neural information processing systems},
  3882--3890.

\bibitem[\protect\citeauthoryear{Ong, Zaki, and
  Goodman}{2015}]{ong2015affective}
Ong, D.~C.; Zaki, J.; and Goodman, N.~D.
\newblock 2015.
\newblock Affective cognition: Exploring lay theories of emotion.
\newblock {\em Cognition} 143:141--162.

\bibitem[\protect\citeauthoryear{Rezende, Mohamed, and
  Wierstra}{2014}]{rezende2014stochastic}
Rezende, D.~J.; Mohamed, S.; and Wierstra, D.
\newblock 2014.
\newblock Stochastic backpropagation and approximate inference in deep
  generative models.
\newblock In {\em International Conference on Machine Learning},  1278--1286.

\bibitem[\protect\citeauthoryear{Wu and Goodman}{2018}]{wu2018multimodal}
Wu, M., and Goodman, N.
\newblock 2018.
\newblock Multimodal generative models for scalable weakly-supervised learning.
\newblock In {\em Advances in Neural Information Processing Systems},
  5575--5585.

\end{thebibliography}


\begin{thebibliography}{1}

\bibitem{wu2018multimodal}
Mike Wu and Noah Goodman.
\newblock Multimodal generative models for scalable weakly-supervised learning.
\newblock In {\em Advances in Neural Information Processing Systems}, pages
  5575--5585, 2018.

\bibitem{krishnan2017structured}
Rahul~G Krishnan, Uri Shalit, and David Sontag.
\newblock Structured inference networks for nonlinear state space models.
\newblock In {\em Thirty-First AAAI Conference on Artificial Intelligence},
  2017.

\bibitem{bowman2015generating}
Samuel~R Bowman, Luke Vilnis, Oriol Vinyals, Andrew~M Dai, Rafal Jozefowicz,
  and Samy Bengio.
\newblock Generating sentences from a continuous space.
\newblock {\em arXiv preprint arXiv:1511.06349}, 2015.

\end{thebibliography}

\end{document}

\maketitle

\appendix
\makeatletter
\renewcommand{\thefigure}{S\arabic{figure}}
\renewcommand{\thetable}{S\arabic{table}}
\makeatletter

\section{Products and Quotients of Gaussian Distributions}

Here we state the closed form solution for the quotient of two products a finite number of Gaussian distributions. We define:
\[g(x) = \prod_{i=1}^k f(x|\mu_i,\Sigma_i) \bigg/ \prod_{j=k+1}^l f(x|\mu_j,\Sigma_j),\]
where each $f(x|\mu_i,\Sigma_i)$ is a multivariate Gaussian with mean $\mu_i$, covariance $\Sigma_i$ and precision $T_i = \Sigma_i^{-1}$. Under the constraint that $\sum_{i=1}^k T_i > \sum_{j=k+1}^l T_j$ element-wise, $g(x)$ is also multivariate Gaussian with mean and covariance:
\begin{align*}
    \mu_g = \frac{\sum_{i=1}^k \mu_i T_i - \sum_{j=k+1}^l \mu_j T_j}{\sum_{i=1}^k T_i - \sum_{j=k+1}^l T_j} \qquad \Sigma_g = \left(\sum_{i=1}^k T_i - \sum_{j=k+1}^l T_j\right)^{-1}
\end{align*}
If the constraint is not satisfied, then $g(x)$ cannot be normalized into a well-defined probability distribution. The reader may refer to \cite{wu2018multimodal} for a modern proof.

\section{Multimodal Bidirectional Training Loss}

Bidirectional factorized variational inference (BFVI) requires training the Multimodal Deep Markov Model (MDMM) to perform both backward filtering and forward smoothing. In addition, the model has to learn how to perform inference given multiple modalities, both jointly and in isolation. As such, we extend the multimodal training paradigm proposed for the MVAE \cite{wu2018multimodal}. For each batch of training data, we compute and minimize the following multimodal bidirectional training loss:
{\small
\begin{align*}
    L = \lambda_\text{filter} \left[L_\text{filter}^{1:M} + \sum_{m=1}^M L_\text{filter}^{m}\right] + \lambda_\text{smooth} \left[L_\text{smooth}^{1:M} + \sum_{m=1}^M L_\text{smooth}^{m}\right] + \lambda_\text{match} L_\text{match}
\end{align*}
}
Here, $\lambda_\text{filter}$, $\lambda_\text{smooth}$ and $\lambda_\text{match}$ are loss multipliers.  $L_\text{filter}^{1:M}$ and $L_\text{smooth}^{1:M}$ are the multimodal ELBO losses for filtering and smoothing respectively:
{\small
\begin{align*}
    L_\text{filter}^{1:M} &= \sum_{t=1}^T \left[ \mathop{\mathbb{E}}_{q(z_t|x_{t:T}^{1:M})} \sum_{m=1}^M \lambda_m \log p(x_t^m|z_t)  - \mathop{\mathbb{E}}_{q(z_{t+1}|x_{t+1:T}^{1:M})} \beta\  \text{KL}\bigg(q(z_t|z_{t+1},x_t^{1:M}) \bigg|\bigg| p(z_t|z_{t+1})\bigg) \right] \\
    L_\text{smooth}^{1:M} &= \sum_{t=1}^T \left[ \mathop{\mathbb{E}}_{q(z_t|x_{1:T}^{1:M})} \sum_{m=1}^M \lambda_m \log p(x_t^m|z_t) - \mathop{\mathbb{E}}_{q(z_{t-1}|x_{1:T}^{1:M})}  \ \ \beta\  \text{KL}\bigg(q(z_t|z_{t-1},x_{t:T}^{1:M}) \bigg|\bigg| p(z_t|z_{t-1})\bigg) \right]
\end{align*}
}
$\lambda_m$ is the reconstruction loss multiplier for modality $m$, and $\beta$ is the loss multiplier for the KL divergence. $L_\text{filter}^m$ and $L_\text{smooth}^m$ are the corresponding unimodal ELBO losses:
{\small
\begin{align*}
    L_\text{filter}^m &= \sum_{t=1}^T \left[ \mathop{\mathbb{E}}_{q(z_t|x_{t:T}^m)} \lambda_m \log p(x_t^m|z_t)  - \mathop{\mathbb{E}}_{q(z_{t+1}|x_{t+1:T}^m)} \beta\  \text{KL}\bigg(q(z_t|z_{t+1},x_t^m) \bigg|\bigg| p(z_t|z_{t+1})\bigg) \right] \\
    L_\text{smooth}^m &= \sum_{t=1}^T \left[ \mathop{\mathbb{E}}_{q(z_t|x_{1:T}^m)} \lambda_m \log p(x_t^m|z_t) - \mathop{\mathbb{E}}_{q(z_{t-1}|x_{1:T}^m)}  \ \beta\  \text{KL}\bigg(q(z_t|z_{t-1},x_{t:T}^m) \bigg|\bigg| p(z_t|z_{t-1})\bigg) \right]
\end{align*}
}
$L_\text{match}$ is the prior matching loss, to ensure that the forward and backward dynamics conform the assumption that $p(z_t)$ is invariant with $t$:
{\small
\[L_\text{match} = \text{KL}\bigg(p(z_t)\bigg|\bigg|\mathbb{E}_{z_{t-1}} p(z_t|z_{t-1})\bigg) + \text{KL}\bigg(p(z_t)\bigg|\bigg|\mathbb{E}_{z_{t+1}} p(z_t|z_{t+1})\bigg)\]
}
To compute $L_\text{match}$, we need to sample particles from $p(z_t)$, introducing $K_p$, the number of prior matching particles, as a hyper-parameter. Similarly, we need to perform backward filtering with sampling to compute the smoothing ELBOs, for which we use $K_b$ backward filtering particles.

$L_\text{filter} := L_\text{filter}^{1:M} + \sum_{m=1}^M L_\text{filter}^{m}$ and $L_\text{smooth}^{1:M} := L_\text{smooth}^{1:M} + \sum_{m=1}^M L_\text{smooth}^{m}$ are minimized together so as to ensure that the model learns accurate backward dynamics. This is done because computation of $L_\text{smooth}$ involves a backward pass which requires sampling under the backward filtering distribution. As such, to accurately approximate $L_\text{smooth}$, the backward filtering distribution has to be reasonably accurate as well, which motivates optimizing $L_\text{filter}$ jointly.

Empirically, we found that optimizing $L_\text{smooth}$ alone led to poor performance for backward extrapolation, and to small gradients for the backward transition parameters, suggesting that the backward pass was too far upstream in the computation graph.
Optimizing $L_\text{filter}$ in addition to $L_\text{smooth}$ rectified this issue with vanishing gradients. While we chose to weight $L_\text{filter}$ and $L_\text{smooth}$ equally, many other training schemes are possible, e.g., placing a high weight on $L_\text{filter}$ initially and then annealing it to zero, such that only $L_\text{smooth}$ (i.e. the correct ELBO if one's goal is to perform only smoothing) is optimized eventually. Such training schemes should be investigated in future work.

\section{Model and Inference Architectures}

\subsection{Transition Functions}

We use a variant of the Gated Transition Function (GTF) from \cite{krishnan2017structured} to parameterize both the forward transition $p(z_t|z_{t-1})$ and backward transition $p(z_t|z_{t-1})$:
\begin{align*}
    g_t &= \text{Sigmoid} \circ \text{Linear}_{|z|\gets|h|} \circ \text{ReLU} \circ \text{Linear}_{|h|\gets|z|} (z_t) \\
    \tilde{\mu}_t &= \text{Linear}_{|z|\gets|h|} \circ \text{ReLU} \circ \text{Linear}_{|h|\gets|z|} (z_t)  \\
    \overline{\mu}_t &= \text{Linear}_{|z|\gets|z|} (z_t) \\
    \mu_t &= g_t \odot \tilde{\mu}_t + (1-g_t) \odot \overline{\mu}_t \\
    \sigma_t &= \text{Softplus} \circ \text{Linear}_{|z|\gets|z|} (\tilde{\mu}_t)
\end{align*}
where Linear$_{b \gets a}$ is a linear  mapping from $a$ to $b$ dimensions, $\circ$ is function composition, and $\odot$ is element-wise multiplication, $|z|$ is the dimension of the latent space (5 for spirals, 256 for videos), and $|h|$ is the dimension of the hidden layer (20 for spirals, 256 for videos). To stabilize training when using BFVI, we found it necessary to add a small constant ($0.001$) to the standard deviation, $\sigma_t$. To be clear, we learn separate networks each for the forward and backward transitions.

\subsection{Encoders and Decoders}

For the Spirals dataset, we used multi-layer perceptrons as encoders and decoders for the $x$ and $y$ data, with $|h|=20$ hidden units. Let $v$ stand for either $x$ or $y$, and recall that for the encoder, we learn a network that parameterizes the quotient $\tilde q(z_t|v_t) = q(z_t|v_t) / p(z_t)$. The architectures are: \\
\begin{minipage}{0.475\textwidth}
\centering
\begin{align*}
    \tilde{q}(z_t|v_t) &\sim \mathcal{N}(\mu_z, \sigma_z)  \\
    h &= \text{ReLU} \circ \text{Linear}_{|h|\gets|1|} (v_t) \\
    \mu_z &= \text{Linear}_{|z|\gets|h|} (h) \\
    \sigma_z &= \text{Softplus} \circ \text{Linear}_{|z|\gets|h|} (h)
\end{align*}
\end{minipage}
\begin{minipage}{0.475\textwidth}
\centering
\begin{align*}
    p(v_t|z_t) &\sim \mathcal{N}(\mu_v, \sigma_v)  \\
    h &= \text{ReLU} \circ \text{Linear}_{|h|\gets|z|} (z_t) \\
    \mu_v &= \text{Linear}_{1\gets|h|} (h) \\
    \sigma_v &= \text{Softplus} \circ \text{Linear}_{1\gets|h|} (h)
\end{align*}
\end{minipage}

For the Weizmann video dataset, we used a variant of the above to work with categorical distributions over the action labels. Again we use $v$ as a placeholder. The architectures are:\\
\begin{minipage}{0.475\textwidth}
\centering
\begin{align*}
    \tilde{q}(z_t|v_t) &\sim \mathcal{N}(\mu_z, \sigma_z)  \\
    e &= \text{ReLU} \circ \text{Embedding}_{|h|} (v_t) \\
    h &= \text{ReLU} \circ \text{Linear}_{|h|\gets|h|} (e) \\
    \mu_z &= \text{Linear}_{|z|\gets|h|} (h) \\
    \sigma_z &= \text{Softplus} \circ \text{Linear}_{|z|\gets|h|} (h)
\end{align*}
\end{minipage}
\begin{minipage}{0.475\textwidth}
\centering
\begin{align*}
    p(v_t|z_t) &\sim \text{Categorical}(p_1, ..., p_k)  \\
    h &= \text{ReLU} \circ \text{Linear}_{|h|\gets|z|} (z_t) \\
    (p_1, ..., p_k) &= \text{Softmax} \circ \text{Linear}_{k\gets|h|} (h)
\end{align*}
\end{minipage}

The number of categories is denoted by $k$ (10 for actions), and we used $|h|=256$. For the encoders and decoders from and to images (video frames or silhouette masks), see Figure \ref{fig:SUPP:img_encdec} below.

\begin{figure}[htb]
    \centering
    \begin{subfigure}{0.475\textwidth}
    \centering
    \includegraphics[width=\textwidth]{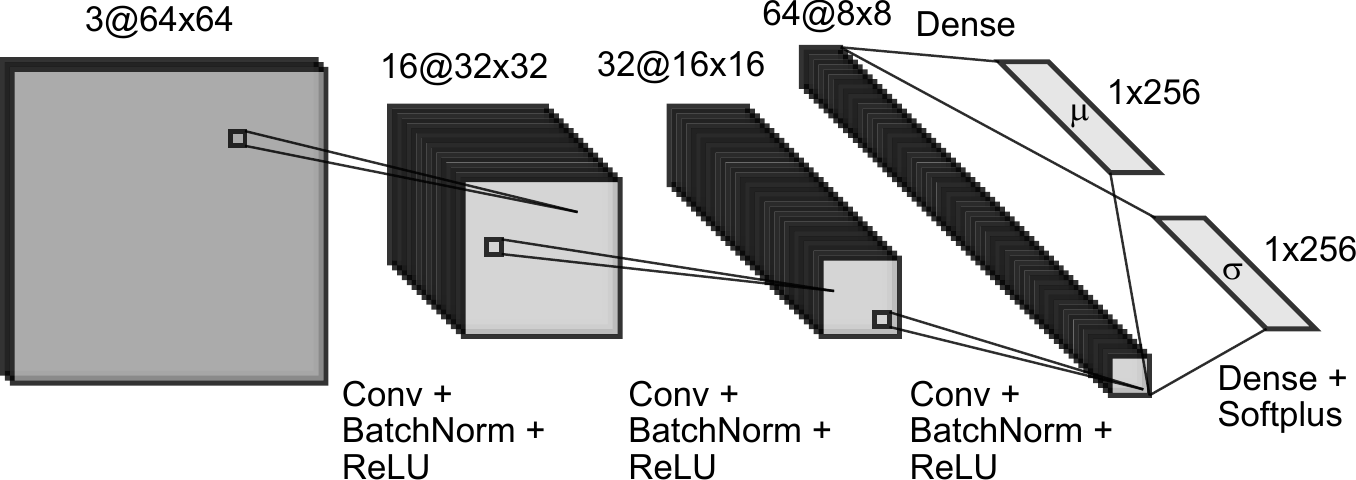}
    \caption{Image Encoder}
    \label{fig:SUPP:img_enc}
    \end{subfigure}
    \begin{subfigure}{0.475\textwidth}
    \centering
    \includegraphics[width=\textwidth]{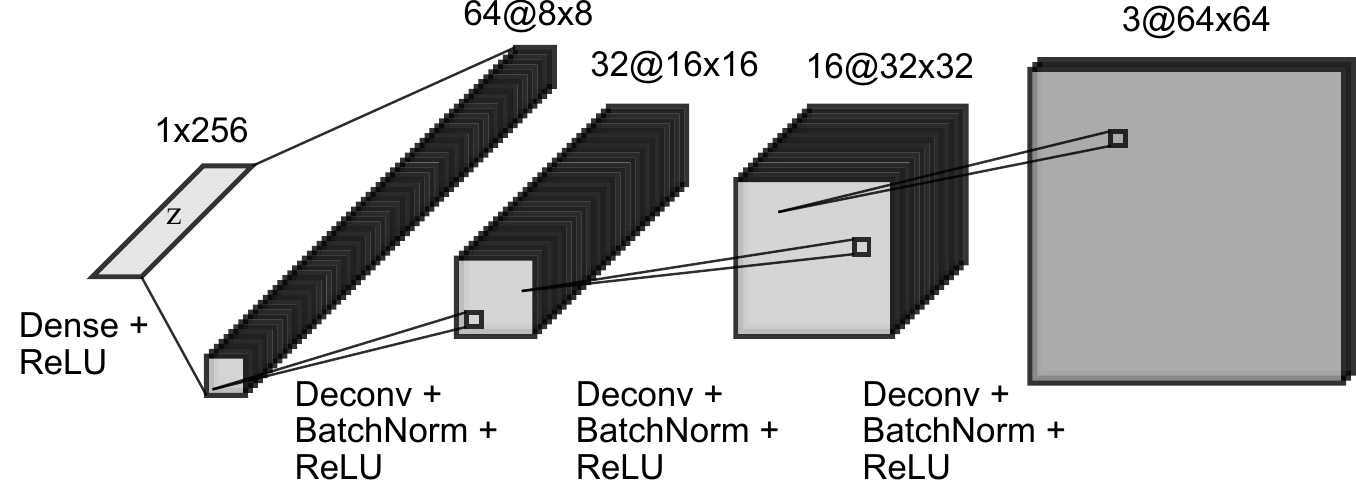}
    \caption{Image Decoder}
    \label{fig:SUPP:img_dec}
    \end{subfigure}
    \caption{Image encoder and decoder architectures. Convolutional layers use 3$\times$3 kernels, de-convolutional layers use 4$\times$4 kernels, both use a stride of $2$ and padding of $1$. Silhouette masks were had 1 input channel instead of 3 input channels.}
    \label{fig:SUPP:img_encdec}
\end{figure}

\subsection{Inference Networks}

The encoder architectures described in the previous section are designed to work with BFVI, and so they directly output distribution parameters $\mu_z$, $\sigma_z$ for the latent state $z_t$. To adapt them to work with the RNN-based structured inference networks (F-Mask, F-Skip, B-Mask, B-Skip), we simply remove the Gaussian output layers from each encoder, using the features $f_t^m$ of the penultimate layer as inputs to the RNN at each time step.

We follow the original DMM implementation \cite{krishnan2017structured} in using Gated Recurrent Units (GRUs) for the structured inference networks. To generalize this to the multimodal case, we have one GRU per modality, which we label $\text{GRU}_m$ for modality $m$. The difference between the Mask and Skip methods is that the former uses zero-masking of missing inputs:
\[h_{t\pm1}^m =
\begin{cases}
\text{GRU}_m(f_t^m, h^t_m) \qquad &\text{if $x_t^m$ is present} \\
\text{GRU}_m(0, h^t_m) \qquad &\text{otherwise}
\end{cases}
\]
whereas the latter skips the update for the hidden state $h_t^m$ of $\text{GRU}_m$ whenever $x_t^m$ is missing:
\[h_{t\pm1}^m =
\begin{cases}
\text{GRU}_m(f_t^m, h^t_m) \qquad &\text{if $x_t^m$ is present} \\
h^t_m \qquad &\text{otherwise}
\end{cases}
\]
Update skipping is the method used in \cite{krishnan2017structured}'s implementation to handle missing data. For the forward RNNs (F-Mask and F-Skip), we take the positive sign in the plus-minus signs above, while for the backward RNNs (B-Mask and B-Skip), we take the negative sign.

In order to combine the information from the RNNs with the previous latent state $z_{t-1}$ to infer the current latent state $z_t$, we use a variant of the combiner function from \cite{krishnan2017structured} that takes in $z_{t-1}$ and the hidden states $h^t_m$ of all modalities as inputs:
\begin{align*}
    \tilde{q}(z_t|z_{t-1}, x_{1:T}) &\sim \mathcal{N}(\mu_z, \sigma_z)  \\
    h &= \text{ReLU} \circ \text{Linear}_{|h|\gets} (z_{t-1}, h_t^1, ..., h_t^M) \\
    \mu_z &= \text{Linear}_{|z|\gets|h|} (h) \\
    \sigma_z &= \text{Softplus} \circ \text{Linear}_{|z|\gets|h|} (h)
\end{align*}
We use $|h|=5$ for the spirals dataset and $|h|=256$ for the videos (same dimensions for the both the hidden layer and the GRU hidden states). Because the above formulation has no direct connection from the current input $x_t$ to the current latent state $z_t$, it seemed like it would hurt performance when we needed $z_t$ to be a good low-dimensional representation of $x_t$. As such, for the video dataset, we used a variant that also takes in the encoded features $f_t^m$ as inputs (if $x_t^m$ is missing, then the corresponding feature $f_t^m$ is zero-masked):
\begin{align*}
    \tilde{q}(z_t|z_{t-1}, x_{1:T}) &\sim \mathcal{N}(\mu_z, \sigma_z)  \\
    h &= \text{ReLU} \circ \text{Linear}_{|h|\gets} (z_{t-1}, h_t^1, ..., h_t^M, f_t^1, ..., f_t^M) \\
    \mu_z &= \text{Linear}_{|z|\gets|h|} (h) \\
    \sigma_z &= \text{Softplus} \circ \text{Linear}_{|z|\gets|h|} (h)
\end{align*}

Table \ref{tab:SUPP:netparams} compares the number of neural network parameters (model networks + inference networks) required for BFVI vs. the RNN-based inference methods. Despite matching hidden layer dimensions, BFVI still uses 2 to 3 times less parameters, because it does not require inference RNNs for each modality, and uses Product-of-Gaussians for fusion instead of a combiner network.

\begin{table}[H]
\small
\centering
\begin{tabular}{@{}lll@{}}
\toprule
\textbf{Method} & \multicolumn{2}{l}{\textbf{Number of parameters}} \\
                & \textit{Spirals}        & \textit{Weizmann}       \\ \midrule
BFVI            & 1854                    & 4542503                 \\
RNN-based       & 7124                    & 7494183                 \\ \bottomrule
\end{tabular}
\vspace{0.05in}
\caption{Number of network parameters for each method.}
\label{tab:SUPP:netparams}
\end{table}

\section{Training Parameters}

\begin{table}[H]
\small
\centering
\begin{tabular}{@{}llll@{}}
\toprule
\textbf{Parameter}           &    & \textbf{Spirals}                  & \textbf{Weizmann}          \\ \midrule
Filtering ELBO mult.    & $\lambda_\text{filter}$ & 0.5                               & ''                         \\
Smoothing ELBO mult.    & $\lambda_\text{smooth}$ & 0.5                               & ''                         \\
Prior matching mult.    & $\lambda_\text{match}$  & 0.01 $\beta$                      & ''                         \\
Reconstruction mult.   & $\lambda_m$        & $x, y$: 1                         & Vid.: 1, Sil.: 1, Act.:10   \\
KL divergence mult.               & $\beta$            & Anneal(0,1) over $E_\beta$ & ''                         \\
Bwd. filtering particles & $K_b$              & 25                                & ''                         \\
Prior matching particles     & $K_p$              & 50                                & ''                         \\
Training epochs              & $E$                  & 500                               & 400                       \\
$\beta$-annealing epochs     & $E_{\beta}$                   & 100                               & 250                       \\
Early stopping               &                    & Yes                               & No                        \\
Batch size                   &                    & 100                               & 25                         \\
Sequence splitting           &                    & No                                & Into 25 time-step segments \\
Optimizer                    &                    & ADAM                              & ''                         \\
Learning rate                &                    & 0.02 (BFVI), 0.01 (Rest)         & $5 \times 10^{-4}$         \\
Weight decay                 &                    & $1 \times 10^{-4}$                &                            \\
Burst deletion rate          &                    & 0.1                               & 0.2                        \\ \bottomrule
\end{tabular}
\vspace{0.05in}
\caption{Training parameters for spirals and Weizmann datasets.}
\label{tab:SUPP:trainparams}
\end{table}

Table \ref{tab:SUPP:trainparams} provides the training parameters we used. Unless specified, parameters were kept constant across inference methods. As in most VAE-like models, we anneal $\beta$, the multiplier for the KL divergence loss from 0 to 1 over time. This incentivizes the model to first find encoders and decoders that reconstruct well, before regularizing the latent space \cite{bowman2015generating}. Since the prior matching multiplier $\lambda_\text{match}$ also serves to regularize the latent space, we tie its value to $\beta$, so that it increases as $\beta$ increases.

To speed up training on the video dataset, we split each input sequence into segments that were 25 time steps long, and trained on those segments. To improve the robustness to missing data when training each methods, we also introduced burst deletion errors --- random contiguous deletions of inputs. For the spirals dataset, we deleted input segments at training time that were $0.1$ long as the original video lengths. For the video dataset, we deleted input segments at training time that were $0.2$ long as the original video lengths. Deletion start points were selected uniformly at random.

We briefly discuss training differences between BFVI and other methods here. For the RNN-based methods, no bidirectional training is involved, so we only minimized the (forward) filtering ELBO (for F-Mask, F-Skip) or the smoothing ELBO (for B-Mask, B-Skip). While we also tried to use the multimodal training paradigm (i.e. minimizing the sum of $L^{1:M}$ and $L^m$) for the RNN-based methods, the effects of this were mixed: On the spirals dataset, performance dropped very sharply with the multimodal paradigm, whereas performance increased on the Weizmann video dataset. As such, the results we report in the main manuscript use the multimodal training paradigm only for the video dataset. For the spirals dataset, we only minimize $L^{1:M}$, with all inputs provided, but not $L^m$, which is computed with only modality $m$ provided as input. Finally, we used a higher learning rate (0.02) for BFVI on the spirals dataset because we noticed slow convergence with the lower rate of 0.01. Increasing the learning rate to 0.02 for the other methods hurt their performance.

\section{Evaluation Details}

When evaluating the models, we estimate the MAP latent sequence $z_{1:T}^* = \arg\max_{z_{1:T}} p(z_{1:T}|x_{1:T})$ by estimating $\hat z_t = \arg\max_{z_{1:T}} q(z_t|\hat z_{t-1},x_{t:T})$ for each $t$ (i.e., we recursively take the mean of the approximate conditional smoothing posterior, $q(z_t|\hat z_{t-1},x_{1:T})$.) For reconstruction, we then decode this inferred latent sequence and take the MLE of the observations, $\hat x_{1:T} = \arg\max_{x_{1:T}} p(x_{1:T}|\hat z_{1:T})$. While this does not infer the exact MAP sequence, it has been standard practice for recurrent latent variable models \cite{krishnan2017structured}, and we find it empirically adequate in lieu of approximate Viterbi sequencing.

In the case of BFVI, we also have to specify number of filtering particles $K_b$ for the backward pass at evaluation time. We used $K_b=200$ for additional stability, though we found that lower values also produced similar results.

To evaluate the performance of BFVI and the next-best method under weakly supervised learning, we performed 10 training runs for each method and each level of data deletion, then reported the average of best 3 runs. This was done in order to exclude outlying runs where training instability emerged, leading to divergence of the loss function. At test time, the inference task for the Spirals dataset was to reconstruct each spiral given that the first and last 25\% of data points were removed, and an additional 50\% of data points were deleted at random. The inference task for the Weizmann dataset was to predict the action labels when only the video modality was provided, and 50\% of the video frames were missing.

\section{Additional Results}

\subsection{Spiral reconstructions}

\begin{figure}[h]
    \centering
    \includegraphics[width=0.9\textwidth]{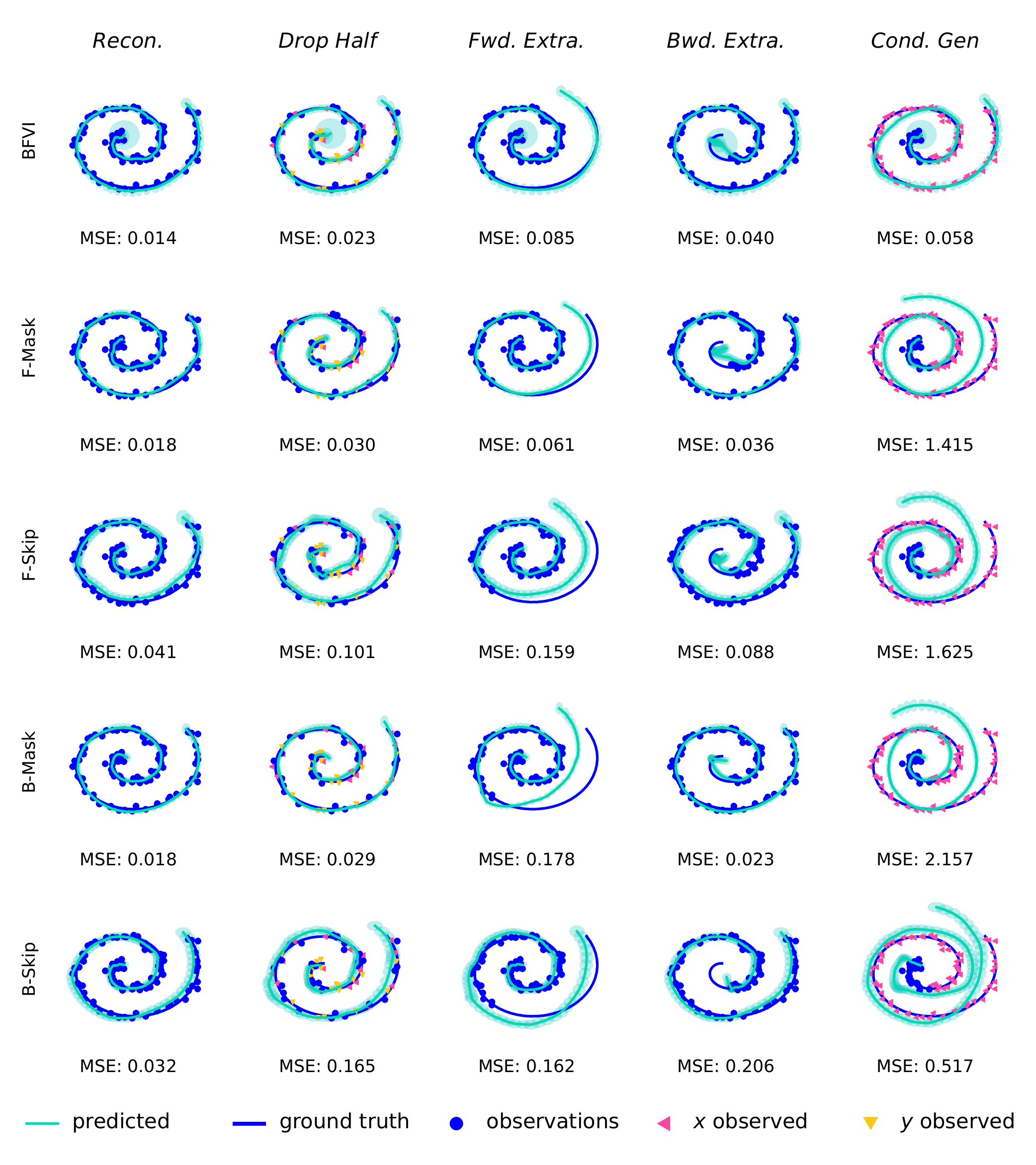}
    \caption{Reconstructions for all 5 spiral inference tasks (columns) for all methods (rows). Our BFVI method does well on all tasks, but especially better than the other methods on the Conditioned Generation task (right-most column), where only the $x$- and first 25\% of the $y$-coordinates are given.}
    \label{fig:SUPP:spirals}
\end{figure}

To supplement the reconstruction figures in the main paper which only compared BFVI to the next-best method, here we show a more complete comparison of all 5 methods on the Spirals dataset (Figure \ref{fig:SUPP:spirals}).

\bibliographystyle{unsrt}
\bibliography{sources}